\documentclass[10pt,twocolumn,letterpaper]{article}

\usepackage{iccv}
\usepackage{times}
\usepackage{epsfig}
\usepackage{graphicx}
\usepackage{amsmath}
\usepackage{amssymb}

\usepackage{booktabs}
\usepackage{inconsolata}
\usepackage{tabu}
\usepackage{color}
\usepackage{tabularx}
\usepackage{pifont}
\usepackage{makecell}
\usepackage{multirow}
\usepackage{algorithm}
\usepackage{algpseudocode}
\usepackage{subcaption}

\usepackage[pagebackref=true,breaklinks=true,letterpaper=true,c olorlinks,bookmarks=false]{hyperref}

\iccvfinalcopy 

\def\iccvPaperID{8700} 

\ificcvfinal\fi

\begin{document}

\title{Models See Hallucinations: Evaluating the Factuality in Video Captioning}

\author{Hui Liu, Xiaojun Wan\\
Wangxuan Institute of Computer Technology, Peking University\\
The MOE Key Laboratory of Computational Linguistics, Peking University\\
{\tt\small \{xinkeliuhui,wanxiaojun\}@pku.edu.cn}
}

\maketitle
\ificcvfinal\thispagestyle{empty}\fi

\begin{abstract}
   Video captioning aims to describe events in a video with natural language. In recent years, many works have focused on improving captioning models' performance. However, like other text generation tasks, it risks introducing factual errors not supported by the input video. These factual errors can seriously affect the quality of the generated text, sometimes making it completely unusable. Although factual consistency has received much research attention in text-to-text tasks (e.g., summarization), it is less studied in the context of vision-based text generation. In this work, we conduct a detailed human evaluation of the factuality in video captioning and collect two annotated factuality datasets. We find that 57.0\% of the model-generated sentences have factual errors, indicating it is a severe problem in this field. However, existing evaluation metrics are mainly based on n-gram matching and show little correlation with human factuality annotation. We further propose a weakly-supervised, model-based factuality metric FactVC, which outperforms previous metrics on factuality evaluation of video captioning. The datasets and metrics will be released to promote future research for video captioning. 
\end{abstract}

\section{Introduction}

\begin{table}[ht]
    \begin{center}
    \begin{tabularx}{\linewidth}{X}
    \toprule
    Video content:\\
    \includegraphics[width=\linewidth]{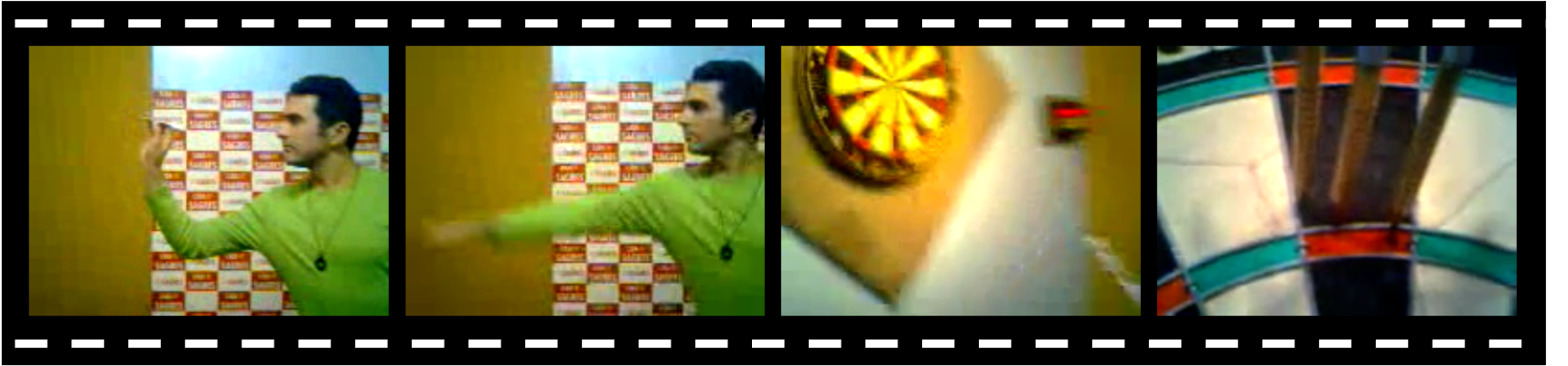}\\ 
    \midrule
    Caption 1:\\
    A \textcolor{red}{\emph{woman}} is throwing darts at a board.\\
    \textcolor{red}{\emph{She}} throws them at a board.\\
    \textcolor{red}{\emph{She jumps off into the distance and smiles}}.\\
    \midrule
    Caption 2:\\
    A man is seen standing in a room and leads into a man \textcolor{red}{\emph{speaking to the camera}}.\\
    The man is throwing darts at a dart board .\\
    The man then throws the dart \textcolor{red}{\emph{board}} and then \textcolor{red}{\emph{goes back to the camera}}.\\
    \midrule
    Caption 3:\\
    A man in a \textcolor{red}{\emph{white}} shirt is standing \textcolor{red}{\emph{at a dart board}}.\\
    He throws a dart at the end.\\
    \bottomrule
  \end{tabularx}
  \end{center}
  \caption{Factual error examples in video captioning. We show a video with three output captions from different captioning models. Factual errors are marked in red italics.}
  \label{tab1}
\end{table}

Video captioning is a challenging cross-modal task that aims to describe videos with natural language sentences. With the rapid development of social media platforms such as Youtube and Tiktok, video plays a more important role in our daily life. Video captioning has received much attention in computer vision and natural language processing communities. Substantial progress has been made to generate descriptions for videos that contain a single event \cite{venugopalan2015sequence, pan2020spatio} or multiple events \cite{zhou2018end, lei2020mart}. However, like other text generation tasks, video captioning models risk introducing factual errors not supported by the input video. Examples are shown in Table \ref{tab1}. This paper defines factual errors (or hallucinations) as follows: a span of caption text that contradicts the video or describes something not appearing in the video. A caption with factual errors can cause misunderstandings of the video content. Sometimes, factual errors make the generated captions completely unusable. 

Factual consistency evaluation has received much research attention in text-to-text tasks, including summarization \cite{maynez2020faithfulness, kryscinski2020evaluating}, knowledge-grounded dialogue \cite{honovich2021q2}, text simplification \cite{devaraj2022evaluating}, and large language models\cite{bang2023multitask}. Nevertheless, it is less studied in vision-to-text tasks, especially video captioning. Therefore, this work focuses on the research gap in the factuality evaluation of video captioning.

Recently, more works have focused on videos with multiple events \cite{song2021towards, wang2021end}, and it may bring more factual errors. So we choose multiple events video captioning for our evaluation. We use ActivityNet Captions \cite{krishna2017dense} and YouCook2 \cite{zhou2018towards} as our video datasets, for they are the most common datasets for this task. Then we carefully select five recent models on each dataset to generate video captions. The models differ in model framework, pretrained features, and input signals. After collecting the videos and captions, we design a factuality annotation protocol and conduct our human annotation. In the end, we obtain two human-annotated factuality datasets ActivityNet-Fact (200 videos, 3,152 sentences) and YouCook2-Fact (100 videos, 3,400 sentences). 

After analyzing the human annotation, we find that factual error (hallucination) is a severe problem in video captioning. To sum up, there are 87.8\% of the paragraphs, 57.0\% of the sentences, and 15.2\% of the words have factual errors. There are different types of factual errors, including person-related errors, action-related errors, object-related errors and so on.

Since hallucination is a severe problem in video captioning, we test to what extent existing automatic evaluation metrics can measure the factuality of video captions. We test commonly used video caption metrics like BLEU\cite{papineni2002bleu}, ROUGE\cite{lin2004rouge}, METEOR\cite{banerjee2005meteor}, CIDEr\cite{vedantam2015cider}, and model-based metrics BERTScore\cite{zhang2019bertscore} and EMScore\cite{shi2022emscore}. We find that most existing metrics correlate poorly with human judgment.

We try to find a better metric to evaluate the factuality in video captioning. Inspired by EMScore\cite{shi2022emscore}, we also leverage the CLIP model to encode video frames and captions. Considering the CLIP model is trained on image-text pairs, it may have a gap transferring to video factuality evaluation. So we automatically construct a training set using text augmentation skills and finetune CLIP on it. Our new metric \textbf{FactVC} (\textbf{Fact}ual consistency for \textbf{V}ideo \textbf{C}aptioning) outperforms all previous metrics, achieving a higher correlation with human factuality annotation.

The main contributions of this work are as follows:
\begin{itemize}
\item We conduct the first thorough factuality evaluation on video captioning. We find that hallucination is a severe problem in this field while existing evaluation metrics can hardly measure it.  
\item We design a factuality annotation protocol and collect two human-annotated factuality datasets for video captioning.
\item We propose a new factuality metric FactVC, which achieves a much higher correlation with human annotation on video captioning, and it can be further transferred to evaluate the factuality of image captioning. 
\item All the datasets, metrics, and code will be released for further research.
\end{itemize}

\section{Related Work}
Factuality evaluation is first proposed in the field of document summarization. Maynez \etal~\cite{maynez2020faithfulness} conducted a human annotation on the XSUM dataset \cite{narayan2018don} and found that more than 70\% of summaries generated by summarization models have factual errors. Other human annotations\cite{wang2020asking, pagnoni2021understanding} reach similar conclusions. To measure the factual consistency, researchers proposed different metrics, which can be roughly divided into Entailment-based metrics \cite{falke2019ranking, kryscinski2020evaluating} and QA-based metrics \cite{durmus2020feqa, wang2020asking}. Inspired by the works in summarization, factuality evaluation is studied for other tasks, including knowledge-grounded dialogue \cite{honovich2021q2}, text simplification \cite{devaraj2022evaluating} and large language models \cite{bang2023multitask}.

For vision-based text generation tasks, the most widely used metrics are based on n-gram matching between reference captions and generated captions, including BLEU\cite{papineni2002bleu}, ROUGE\cite{lin2004rouge}, METEOR\cite{banerjee2005meteor}, and CIDEr\cite{vedantam2015cider}. They are sensitive to lexical variation and cannot match deeper semantics between captions and vision inputs. Recently, there are model-based metrics such as BERTScore\cite{zhang2019bertscore}, CLIPScore\cite{hessel2021clipscore}, EMScore\cite{shi2022emscore}. They leverage large-scale pretrained models to compute a matching score, even without requiring reference captions. 

Little work pays attention to the hallucination problem in vision-based text generation. CHAIR \cite{rohrbach2018object} proposes an image relevance metric to evaluate the object hallucination in image captioning. However, they restrict their evaluation to 80 MSCOCO objects and a single type of hallucination. EMScore\cite{shi2022emscore} is designed for the overall evaluation of video captioning, and it also shows the potential to identify hallucinating captions. However, it is tested on a non-real dataset and is not specifically designed for factuality evaluation.

\section{Human Annotation}
Considering there does not exist factuality annotation of video captioning, we decide to construct our own datasets. We use ActivityNet Captions \cite{krishna2017dense} and YouCook2 \cite{zhou2018towards} as our source video datasets and select five recent models for each dataset to generate captions. Then we design a factuality annotation protocol and conduct our human annotation. 

\subsection{Datasets}
ActivityNet Captions\cite{krishna2017dense} contains 20k untrimmed videos of various human activities. On average, each video lasts 120s and has 3.65 human-annotated sentences. Previous works\cite{lei2020mart,ging2020coot,song2021towards} report results on the ae-test split (2,457 videos). In this work, we randomly sample 200 videos from the ae-test split for human annotation. YouCook2 \cite{zhou2018towards} contains 2,000 long untrimmed videos from 89 cooking recipes. On average, each video lasts 320s and has 7.70 human-annotated sentences. Previous works\cite{lei2020mart,ging2020coot,luo2020univl} report results on the val split (457 videos). We randomly sample 100 videos from the val split for human annotation.

\subsection{Captioning Models}
\label{sec:models}

We select five recent captioning models for each dataset and obtain the output captions on the sampled videos. For ActivityNet Captions, the selected models include: \textbf{MART} \cite{lei2020mart}: a recurrent transformer model that uses a memory module to augment the transformer architecture; \textbf{COOT} \cite{ging2020coot}: it first hierarchically trains video-text representation on the ActivityNet dataset, and then uses the representation to train MART model; \textbf{PDVC-gt} \cite{wang2021end}: it uses parallel decoding to generate dense video captioning. The above three models need the human-annotated event timestamps to generate each sentence; \textbf{PDVC-pred} \cite{wang2021end}: similar to PDVC-gt, but it uses predicted event timestamps to generate sentences; \textbf{Song} \cite{song2021towards}: it eschews the event detection stage and directly generates paragraphs for untrimmed videos. For YouCook2, the selected models include: \textbf{VTrans}\cite{zhou2018end}: a transformer architecture for video captioning; \textbf{MART}\cite{lei2020mart}; \textbf{COOT}\cite{ging2020coot}; \textbf{COOT-100m}\cite{ging2020coot}: it uses the pretrained features on the large-scale Howto100M dataset \cite{miech2019howto100m} to train COOT model; \textbf{UniVL}\cite{luo2020univl}: a unified video and language pretraining model pretrained on Howto100M dataset. The above models are different in model framework, input signals, pretrained features, and pretraining scales, as shown in Table \ref{tab2}.

\begin{table}[htbp]
    \begin{center}
    \begin{tabular}{l|cccc}
    \toprule
    Model & \makecell[c]{Frame-\\work} & \makecell[c]{Time-\\stamps} & \makecell[c]{Pre-\\train} & \makecell[c]{Large\\Scale} \\
    \midrule
    VTrans\cite{zhou2018end} & VTrans & \ding{51} & \ding{55} & \ding{55}\\
    MART\cite{lei2020mart} & MART & \ding{51} & \ding{55} & \ding{55}\\
    COOT\cite{ging2020coot} & MART & \ding{51} & \ding{51} & \ding{55}\\
    COOT-100m\cite{ging2020coot} & MART & \ding{51} & \ding{51} & \ding{51}\\
    UniVL\cite{luo2020univl} & UniVL & \ding{51} & \ding{51} & \ding{51}\\
    PDVC-gt\cite{wang2021end} & PDVC & \ding{51} & \ding{51} & \ding{55}\\
    PDVC-pred\cite{wang2021end} & PDVC & \ding{55} & \ding{51} & \ding{55}\\
    Song\cite{song2021towards} & Song & \ding{55} & \ding{55} & \ding{55}\\
    \bottomrule
    \end{tabular}
    \end{center}
    \caption{Comparison between selected models on Framework (model framework), Timestamps (whether use human-annotated timestamps), Pretrain (whether pretrain features on video dataset), Large Scale (whether use large-scale pretraining).}
    \label{tab2}
\end{table}

\subsection{Annotation Protocol}
Before conducting human annotation, we design an annotation protocol to instruct annotators on how to measure and label the factuality of video captions. For the factuality annotation in summarization, annotators often give a binary label 0/1 for each summary sentence, indicating whether the sentence is factual or not \cite{maynez2020faithfulness, wang2020asking, kryscinski2020evaluating}. However, the video captions have hierarchical structures (paragraph-sentence-word), and we want to obtain the factuality annotation for different granularity. 

So we design a new annotation protocol. Annotators are asked to give three levels of factuality annotation for each video caption. \textbf{Paragraph-level}: For each paragraph, annotators need to give a factuality Likert scale from 1 to 5, where 1 means the paragraph has many severe factual errors, and 5 means there are no obvious factual errors; \textbf{Sentence-level}: For each sentence, annotators give a label 1 if it has factual errors else label 0; \textbf{Word-level}: Within each sentence, annotators need to mark phrases and words that have factual errors. Another critical issue is that we focus on whether the caption has factual errors given the video (Precision) and do not care whether the caption describes the video completely (Recall). Please refer to Appendix \ref{sec:annotation} for the complete annotation protocol and examples.

\begin{table*}[ht]
    \begin{center}
    \begin{tabular}{c|c|c|c|c}
    \toprule
    Category & Description & \multicolumn{2}{c|}{Example} & Ratio\\
    \midrule
    Person & \makecell[c]{Person-related factual errors,\\including gender, age,\\ pronoun errors} & \makecell[c]{\includegraphics[width=0.11\textwidth]{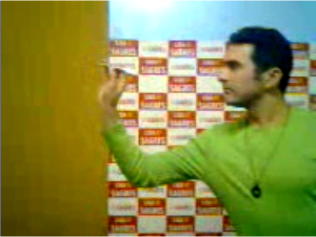}} & \makecell[c]{A \textcolor{red}{\emph{woman}} is throwing\\ darts at a board.} & 25.6\%\\
    \midrule
    Action & \makecell[c]{Action-related factual errors,\\ the action is not consistent\\ with the video} & \makecell[c]{\includegraphics[width=0.11\textwidth]{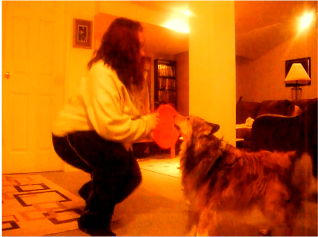}} & \makecell[c]{The woman then begins\\\textcolor{red}{\emph{dancing}} with the dog...} & 38.3\%\\
    \midrule
    Object & \makecell[c]{Object-related factual errors,\\ the object is not consistent\\ with the video} & \makecell[c]{\includegraphics[width=0.11\textwidth]{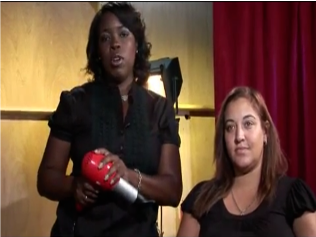}} & \makecell[c]{She then shows off a\\ \textcolor{red}{\emph{rag}} and speaking to\\ the camera.} & 19.6\%\\
    \midrule
    Adjective & \makecell[c]{Adjective-related factual\\errors, such as color,\\ numerical errors} & \makecell[c]{\includegraphics[width=0.11\textwidth]{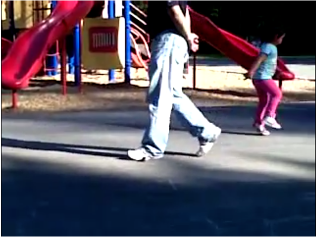}} & \makecell[c]{A person in a \textcolor{red}{\emph{red}} shirt\\ is walking towards the\\ camera.} & 6.6\%\\
    \midrule
    \makecell[c]{Poor\\Generation} & \makecell[c]{The sentence is poorly\\ generated so that it\\ contains factual errors} & \makecell[c]{\includegraphics[width=0.11\textwidth]{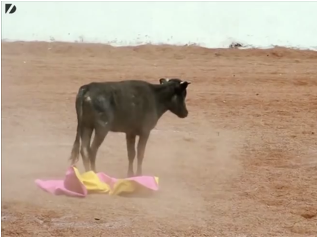}} & \makecell[c]{the bull is \textcolor{red}{\emph{UNK}} and the\\ bull is \textcolor{red}{\emph{UNK}} .} & 5.5\%\\
    \midrule
    Other & \makecell[c]{Other factual errors,\\ including preposition errors,\\ relation errors, etc.} & \makecell[c]{\includegraphics[width=0.11\textwidth]{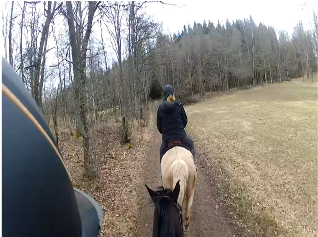}} & \makecell[c]{the person is riding the\\horses \textcolor{red}{\emph{in the air}}.} & 4.4\%\\
    \bottomrule
    \end{tabular}
    \end{center}
    \caption{Typology of factual errors in video captioning. Examples of each category are shown with a related video frame. Factual errors are marked in red italics. Ratios are shown on the ActivityNet-Fact dataset.}
    \label{tab3}
\end{table*}

\subsection{Annotation Procedure}
According to previous works \cite{kryscinski2020evaluating, wang2020asking}, the inter-annotator agreement through the crowdsourcing platform is relatively low. To make our annotation more reliable, we hire three graduate students as our annotators. We provide them with a detailed instruction document and several annotation examples so that they can fully understand the annotation protocol. The annotations are checked multiple times during the annotation process. If there are any problems, we give suggestions to help annotators improve the annotations. The annotations will be adopted only when an annotator completes all videos and passes every check, ensuring the annotations' quality and consistency. We collect three annotations for each video caption and combine them to get the final annotation. For paragraph-level annotation, if there exists a majority score, we use it as the final score; otherwise, we use the median score. For sentence-level and word-level annotation, we use the majority label as the final label. 

We quantified the degree of inter-annotator agreement using Krippendorff's alpha coefficient \cite{krippendorff2011computing}. On the ActivityNet dataset, the inter-annotator interval metrics are 0.741, 0.647, and 0.564 for paragraph-level, sentence-level, and word-level annotations respectively. On the YouCook2 dataset, the inter-annotator interval metrics are 0.783, 0.766, 0.688 for paragraph-level, sentence-level, and word-level annotations respectively. The metrics show a substantial agreement between annotators. The agreement for word-level annotation is relatively low because it has more uncertainty and ambiguity.

\section{Annotation Analysis}

\subsection{Datasets Statistics}
Based on sampled ActivityNet and YouCook2 videos, we collect two annotated factuality datasets ActivityNet-Fact and YouCook2-Fact. The ActivityNet-Fact dataset contains 1,000 paragraphs, 3,152 sentences, and 40,461 words, among which 82.7\% of the paragraphs, 52.7\% of the sentences, and 14.0\% of the words have factual errors. The YouCook2-Fact dataset contains 500 paragraphs, 3,400 sentences, and 24,903 words, among which 98\% of the paragraphs, 60.9\% of the sentences, and 17.1\% of the words have factual errors. This indicates that factual error is a severe problem in video captioning and should attract more research attention.

\begin{table*}[ht]
\begin{center}
\begin{tabular}{c|l|cccc|ccc}
\toprule
\multirow{2}{*}{Datasets} & \multirow{2}{*}{Models} & \multicolumn{4}{c}{Automatic Metrics} & \multicolumn{3}{|c}{Factuality Annotation}\\
\cmidrule{3-9}
& & Bleu4 & METEOR & Rouge-L & CIDEr & Paragraph & Sentence & Word\\
\midrule
\multirow{5}{*}{\makecell[c]{ActivityNet-\\Fact}} & MART & 10.48 & 15.61 & 30.52 & 24.39 & 3.08 & 0.413 & 0.816\\
& COOT & 11.38 & 16.26 & \textbf{31.80} & 27.23 & 3.31 & 0.446 & 0.851\\
& PDVC-gt & \textbf{12.83} & \textbf{16.96} & 28.35 & \textbf{32.32} & \textbf{3.54} & \textbf{0.553} & \textbf{0.882}\\
& PDVC-pred & 10.66 & \underline{16.43} & \underline{31.42} & 19.29 & \underline{3.52} & \underline{0.513} & \underline{0.881}\\
& Song & \underline{11.87} & 15.71 & 30.63 & \underline{31.19} & 3.50 & 0.443 & 0.864\\
\midrule
\multirow{5}{*}{\makecell[c]{YouCook2-\\Fact}} & VTrans & 7.26 & 15.21 & 31.49 & 34.40 & 1.96 & 0.237 & 0.778\\
& MART & 7.66 & 15.44 & 31.87 & 37.32 & 2.00 & 0.261 & 0.774\\
& COOT & 9.70 & 17.82 & 34.51 & 59.39 & 2.57 & 0.380 & 0.833\\
& COOT-100m & \underline{11.42} & \underline{19.25} & \underline{37.65} & \underline{62.41} & \underline{3.01} & \underline{0.490} & \underline{0.870}\\
& UniVL & \textbf{12.96} & \textbf{19.86} & \textbf{41.22} & \textbf{69.20} & \textbf{3.24} & \textbf{0.565} & \textbf{0.892}\\
\bottomrule
\end{tabular}
\end{center}
\caption{Model performance on automatic metrics and Factuality Annotation. We report the average paragraph score and the ratio of factual sentences and words for factuality annotation. The two highest-rated models are in bold and underlined respectively.}
\label{tab4}
\end{table*}

\subsection{Factual Error Type Analysis}
FRANK\cite{pagnoni2021understanding} propose a typology of factual errors towards summarization and annotate their data with specific factual errors. Although we do not annotate the factual error type during human annotation, we conduct a post-analysis with our annotated datasets. We collect phrases/words marked as factual errors appearing at least twice and classify them into different error categories. The results are shown in Table \ref{tab3}. We can see that the factual errors in video captioning are various. For ActivityNet-Fact, the most common factual error categories are Person, Action, and Object, which count for 83.5\% of the total factual errors. For YouCook2-Fact, Object is the dominant category, which counts for 92.7\% of the total factual errors.

\subsection{Model Evaluation}
We evaluate the selected models using automatic metrics and our factuality annotation. The results are shown in Table \ref{tab4}. On the ActivityNet-Fact dataset, the PDVC-gt model performs best on both automatic metrics (except for Rouge-L) and factuality annotation. On YouCook2-Fact dataset, the UniVL model performs best. Combing the model characteristics in Table \ref{tab2}, we can draw several conclusions: 1) Pretraining (especially large-scale pretraining) can improve both the automatic metrics and factuality (compare MART vs. COOT, they use the same model framework but with different features). 2) The timestamps information is helpful in improving automatic metrics but less helpful in improving factuality (compare PDVC-gt with PDVC-pred, PDVC-gt use annotated event timestamps). 3) Model framework is the most important factor for factuality. In addition, Table \ref{tab4} shows that the automatic metrics are not always consistent with factuality annotation.

\section{Metric Analysis}

\begin{table}[htbp]
\begin{center}
\begin{tabular}{l|c|ccc}
\toprule
Metric & ref & Para & Sent & Word\\
\midrule
Bleu1 & T & 0.161 & 0.133 & 0.195\\
Bleu2 & T & 0.204 & 0.184 & 0.238\\
Bleu3 & T & 0.199 & 0.194 & 0.222\\
Bleu4 & T & 0.178 & 0.173 & 0.174\\
METEOR & T & 0.204 & 0.196 & 0.229\\
Rouge-L & T & 0.170 & 0.151 & 0.185\\
CIDEr & T & 0.151 & 0.141 & 0.133\\
\midrule
BERTScore & T & 0.243 & 0.196 & 0.198\\
EMScore & V & 0.305 & 0.242 & 0.341\\
EMScore & T & 0.452 & 0.389 & 0.447\\
EMScore & VT & 0.458 & 0.388 & 0.464\\
\bottomrule
\end{tabular}
\end{center}
\caption{Pearson correlation between automatic evaluation metrics and human annotation on the ActivityNet-Fact dataset. "ref" means the metric reference is human-written caption (V), input video (V), or both (VT).}
\label{tab5}
\end{table}

\begin{table}[htbp]
\begin{center}
\begin{tabular}{l|c|ccc}
\toprule
Metric & ref & Para & Sent & Word\\
\midrule
Bleu1 & T & 0.276 & 0.238 & 0.301\\
Bleu2 & T & 0.304 & 0.298 & 0.337\\
Bleu3 & T & 0.247 & 0.271 & 0.301\\
Bleu4 & T & 0.197 & 0.237 & 0.250\\
METEOR & T & 0.411 & 0.371 & 0.415\\
Rouge-L & T & 0.361 & 0.333 & 0.372\\
CIDEr & T & 0.150 & 0.123 & 0.179\\
\midrule
BERTScore & T & 0.426 & 0.403 & 0.434\\
EMScore & V & 0.346 & 0.372 & 0.350\\
EMScore & T & 0.524 & 0.501 & 0.537\\
EMScore & VT & 0.543 & 0.530 & 0.555\\
\bottomrule
\end{tabular}
\end{center}
\caption{Pearson correlation between automatic evaluation metrics and human annotation on the YouCook2-Fact dataset.}
\label{tab6}
\end{table}

Now that factual errors broadly exist in video captions, we want to know to what extent existing metrics can measure the factuality of video captions. We test the correlation between automatic metrics and human annotation (for sentence/word-level annotation, we use the ratio of factual sentences and words as annotation score). We test model-free metrics BLEU\cite{papineni2002bleu}, ROUGE\cite{lin2004rouge}, METEOR\cite{banerjee2005meteor}, CIDEr\cite{vedantam2015cider}, and model-based metrics BERTScore\cite{zhang2019bertscore} and recently proposed EMScore\cite{shi2022emscore} \footnote{For EMScore, we use the ViT-B/16 CLIP model, which performs better than the default ViT-B/32 CLIP model.}. The results are shown in Table \ref{tab5} and Table \ref{tab6}. To our surprise, the most commonly used metrics for video captioning, such as Bleu4 and CIDEr, correlate poorly with factuality annotation. METEOR and Bleu2  perform relatively better but still show a weak correlation with factuality annotation. As for model-based metrics, BERTScore shows little superior to METEOR on two datasets, indicating that just introducing large-scale text-pretrained model is not enough. The recently proposed EMScore, which leverages the image-text pretrained model CLIP\cite{radford2021learning}, shows a higher correlation with human annotation. In addition, it can evaluate video captions using the input videos, with no need for human-written captions. We attribute the advantage of EMScore to large-scale cross-modal pretraining.

\section{FactVC Metric}
Although EMScore achieves a good correlation with human factuality annotation, it has two drawbacks: 1) EMScore uses the pretrained model CLIP, which is trained on image-text pairs from the Internet, and it may not transfer well to the video captioning data; 2) EMScore is designed for evaluating the overall quality of the video caption, not specifically designed for factuality evaluation. As a result, we propose a new metric \textbf{FactVC} (\textbf{Fact}ual consistency for \textbf{V}ideo \textbf{C}aptioning). We first automatically construct a factuality training set using text augmentation skills and then use it to finetune the CLIP model. We also improve the calculation of the similarity score so that it is more suitable for factuality evaluation.

\subsection{Training Data}
Collecting a large-scale training dataset through human annotation is expensive and time-consuming. Inspired by \cite{kryscinski2020evaluating, gokhale2022semantically}, we decide to construct our training set automatically using text augmentation skills. Given a video $V$ together with a human-annotated caption sentence $T$, we use a set of text transformation functions to augment the dataset. The transformation functions include positive transformations ($\mathcal{T}^{+}$) which ensure the new sentence is factually correct and negative transformations ($\mathcal{T}^{-}$) which introduce factual errors into the sentence.

The positive transformations include: 1)Paraphrasing: we generate paraphrases using the back-translation method. We use the Google Translation API \footnote{\url{https://translate.google.com/}} and use German and French as middle language; 2) Simplification: we use a tool \footnote{\url{https://github.com/garain/Sentence-Simplification}} to simplify complex and compound sentences into simple sentences.

The negative transformations include: 1) Person Swap: we design a set of rules to change person words' gender, age, and pronoun; 2) Action Swap: we collect a common action set and apply deletion and insertion; 3) Object Swap: we collect a common object set and apply object substitution; 4) Adjective Swap: we swap adjectives (color, numerical words, etc.) in original sentences; 5) Poor Generation: we simulate the poor generation sentences by inserting "UNK" word and redundancy phrases. We design the negative transformations according to the factual errors shown in Table \ref{tab3}.

We first apply positive transformations to obtain positive sentences and then apply negative transformations to them to get negative sentences. Finally, we collect a set of data samples $(V, T^{+}, T^{-})$, where $V$ means the input video, $T^{+}$ means a fact-consistent sentence, and $T^{-}$ means the corresponding fact-inconsistent sentence. A detailed description of the data generation process is in Appendix \ref{sec:data_gen}.

\subsection{CLIP Finetuning}
We only finetune the projection layers of the pretrained CLIP model. Given a batch of data $\{(V_i, T_{i}^{+}, T_{i}^{-})\}_{i=1}^{B}$ with a batch size of $B$, we first use the CLIP model to calculate the similarities between each video and text:
\begin{align}
    s_{i,j}^{+} = cos\left( \frac{1}{|V_i|}\sum_{k=1}^{|V_i|}E_v(f_{ik}), E_t(T_{j}^{+})\right)
\end{align}
where $s_{i,j}^{+}$ means the similarity score between $V_i$ and $T_j^+$, $cos$ means cosine similarity, $f_{ik}$ is the $k$-th frame of video $V_i$, $|V_i|$ is the sampled frame number, $E_v$ and $E_t$ are the vision encoder and text encoder of CLIP model. The similarity score $s_{i,j}^{-}$ between $V_i$ and $T_{i}^{-}$ is computed similarly.

Then we finetune the CLIP model using the following loss function:
\begin{align}
    \mathcal{L}_{coarse} &= - \sum_{i=1}^{B} \frac{exp(s_{i,i}^{+})}{\sum_{j=1}^{B}(exp(s_{i,j}^{+}))}\\
    \mathcal{L}_{fine} =& \sum_{i=1}^{B} max(0, M - s_{i,i}^{+} + s_{i,i}^{-})\label{eq3}\\
    \mathcal{L} &= \mathcal{L}_{coarse} + \lambda \mathcal{L}_{fine}\label{eq4}
\end{align}
where $\mathcal{L}_{coarse}$ is a cross-entropy loss to learn whether the video content and text are matched, $\mathcal{L}_{fine}$ is a hinge loss to learn to assign a higher score to the fact-consistent text. $M$ and $\lambda$ are hyper-parameters.

\begin{table*}[ht]
\begin{center}
\begin{tabular}{c|l|ccc|ccc|ccc}
\toprule
\multirow{2}{*}{Datasets} & \multirow{2}{*}{Metrics} & \multicolumn{3}{c}{Video as ref} & \multicolumn{3}{|c}{Text as ref} & \multicolumn{3}{|c}{Video \& Text as ref}\\
\cmidrule{3-11}
& & Para & Sent & Word & Para & Sent & Word & Para & Sent & Word\\
\midrule
\multirow{6}{*}{\makecell[c]{ActivityNet-\\Fact}}& METEOR & - & - & - & 0.204 & 0.196 & 0.229 & - & - & -\\
& BERTScore & - & - & - & 0.243 & 0.196 & 0.198 & - & - & -\\
& Entailment & 0.295 & 0.319 & 0.304 & 0.238 & 0.321 & 0.241 & 0.299 & 0.333 & 0.305\\
& EMScore (ViT-B/32) & 0.253 & 0.190 & 0.300 & 0.425 & 0.356 & 0.432 & 0.427 & 0.352 & 0.446\\
& EMScore (ViT-B/16) & 0.305 & 0.242 & 0.341 & 0.452 & 0.389 & 0.447 & 0.458 & 0.388 & 0.464\\
\cmidrule{2-11}
& FactVC & \textbf{0.462} & \textbf{0.371} & \textbf{0.480} & \textbf{0.511} & \textbf{0.438} & \textbf{0.498} & \textbf{0.551} & \textbf{0.465} & \textbf{0.545}\\
\midrule
\multirow{6}{*}{\makecell[c]{YouCook2-\\Fact}}& METEOR & - & - & - & 0.411 & 0.371 & 0.415 & - & - & -\\
& BERTScore & - & - & - & 0.426 & 0.403 & 0.434 & - & - & -\\
& Entailment & 0.287 & 0.306 & 0.285 & 0.420 & 0.465 & 0.423 & 0.431 & 0.474 & 0.432\\
& EMScore (ViT-B/32) & 0.337 & 0.353 & 0.361 & 0.518 & 0.482 & 0.523 & 0.543 & 0.514 & 0.553\\
& EMScore (ViT-B/16) & 0.346 & 0.372 & 0.350 & 0.524 & 0.501 & 0.537 & 0.543 & 0.530 & 0.555\\
\cmidrule{2-11}
& FactVC & \textbf{0.408} & \textbf{0.410} & \textbf{0.420} & \textbf{0.584} & \textbf{0.558} & \textbf{0.592} & \textbf{0.606} & \textbf{0.583} & \textbf{0.615}\\
\bottomrule
\end{tabular}
\end{center}
\caption{Pearson correlation between automatic metrics and human factuality annotation on two datasets. We test each metric in three settings: video as the reference, text as the reference, video and text as the reference. METEOR and BERTScore only work with text as the reference. Metrics with worse correlation are omitted.}
\label{tab7}
\end{table*}

\begin{table*}[ht]
\begin{center}
\begin{tabular}{l|ccc|ccc|ccc}
\toprule
\multirow{2}{*}{Metrics} & \multicolumn{3}{c}{Video as ref} & \multicolumn{3}{|c}{Text as ref} & \multicolumn{3}{|c}{Video \& Text as ref}\\
\cmidrule{2-10}
& Para & Sent & Word & Para & Sent & Word & Para & Sent & Word\\
\midrule
FactVC & \textbf{0.462} & \textbf{0.371} & \textbf{0.480} & \textbf{0.511} & \textbf{0.438} & \textbf{0.498} & \textbf{0.551} & \textbf{0.465} & \textbf{0.545}\\
\midrule
FactVC(no finetune) & 0.349 & 0.281 & 0.388 & 0.497 & 0.425 & 0.483 & 0.512 & 0.433 & 0.510\\
FactVC($\mathcal{L}_{coarse}$) & 0.427 & 0.348 & 0.466 & 0.508 & 0.435 & 0.492 & 0.537 & 0.456 & 0.532\\
FactVC($\mathcal{L}_{fine}$) & 0.239 & 0.166 & 0.221 & 0.446 & 0.394 & 0.456 & 0.406 & 0.336 & 0.404\\
FactVC(F-value) & 0.442 & 0.352 & 0.454 & 0.474 & 0.403 & 0.472 & 0.512 & 0.428 & 0.514\\
FactVC($\alpha = 0.5$) & 0.444 & 0.366 & 0.458 & 0.485 & 0.423 & 0.474 & 0.523 & 0.450 & 0.518\\
\bottomrule
\end{tabular}
\end{center}
\caption{Ablation study on the ActivityNet-Fact dataset. The pearson correlation between each metric and human factuality annotation.}
\label{tab8}
\end{table*}

\subsection{Score Calculation}
With the finetuned CLIP model, we can calculate the factuality score FactVC as follows:
\begin{align}
    FactVC(T, V) = (1 - \alpha) S(T, V)_{c} + \alpha S(T, V)_{f}^{p} \label{eq5}\\
    S(T, V)_{c} = cos( \frac{1}{|V|}\sum_{k=1}^{|V|}E_v(f_{k}), E_t(T))\\
    S(T, V)_{f}^{p} = \frac{1}{|T|} \sum_{x_j \in T} \max_{f_i \in V} cos(E_v(f_{i}), E_t(x_j))
\end{align}
where $\alpha$ is a balance factor, $S(T, V)_{c}$ is the coarse-grained similarity score between video $V$ and sentence $T$. $S(T, V)_{f}^{p}$ is the precision-based fine-grained similarity score computed between each frame $f$ and each word $x$. Please refer to \cite{shi2022emscore} for more technical details. Similar to EMScore, FactVC can use video $V$, human-written caption $T^{*}$, or both $(V, T^{*})$ as reference.

We improve the calculation of FactVC towards EMScore in two aspects: 1) we use the precision-based score instead of the F-value-based score, for factuality is more related to the precision of video captions; 2) we introduce a parameter $\alpha$ to balance the coarse-grained score and the fine-grained score, we set it to 0.75 to favor more on fine-grained score.

\section{Experiments}

\subsection{Comparison with Other Metrics}
We compare FactVC metric to other automatic metrics including Bleu\cite{papineni2002bleu}, ROUGE\cite{lin2004rouge}, CIDEr\cite{vedantam2015cider}, METEOR\cite{banerjee2005meteor}, BERTScore\cite{zhang2019bertscore}, Entailment\cite{kryscinski2020evaluating}\footnote{Similar to FactCC\cite{kryscinski2020evaluating}, we use our training set to train an entailment model, and use the entailment probability as the factual score.}, EMScore\cite{shi2022emscore}. 

The results are shown in Table \ref{tab7}. We omit the metrics with worse correlation here (Bleu, ROUGE, CIDEr), and you can check them in Tables \ref{tab5} and \ref{tab6}. From the table, EMScore performs better than METEOR, BERTScore, and the Entailment-based metric, indicating the usefulness of pretrained model CLIP. However, EMScore performs relatively poorly using video as the reference. Our FactVC metric, on the other hand, shows a much better performance in this setting. Compared to other metrics, FactVC shows the highest correlation with human annotation in all settings.

\subsection{Ablation Study}
We conduct an ablation study to test the effectiveness of each component in FactVC. The results are shown in Table \ref{tab8}. FactVC(no finetune) removes the finetuning process and shows an obvious performance degradation. FactVC($\mathcal{L}_{coarse}$) only uses $\mathcal{L}_{coarse}$ to finetune CLIP and FactVC($\mathcal{L}_{fine}$) only uses $\mathcal{L}_{fine}$ to finetune CLIP. From the results, we find that $\mathcal{L}_{coarse}$ can ensure stable finetuning, only using $\mathcal{L}_{fine}$ is not a good choice, but it can help video encoding together with $\mathcal{L}_{coarse}$. FactVC(F-value) uses the F-value-based score instead of the precision-based score, showing a performance degradation. This proves that factual consistency is more related to the precision of video captions. FactVC($\alpha=0.5$) sets $\alpha$ to 0.5 in eq (\ref{eq5}) and it is inferior to FactVC with $\alpha=0.75$. This shows that the fine-grained score is more important in Factuality evaluation.

\subsection{Experiments on ActivityNet-FOIL}

\begin{table}[htbp]
\begin{center}
\begin{tabular}{lc|lc}
\toprule
Metric & Acc(\%) & Metric & Acc(\%)\\
\midrule
BLEU1 & 60.1 & EMScore$_f$ & 90.3\\
BLEU4 & 66.1 & EMScore & 89.5\\
Rouge-L & 56.7 & FactVC & 91.0\\
METEOR & 72.9 & EMScore$^{*}_f$ & 93.0\\
CIDEr & 77.9 & EMScore$^{*}$ & 92.4\\
BERTScore & 86.7 & FactVC$^{*}$ & 94.3\\
\bottomrule
\end{tabular}
\end{center}
\caption{Pairwise ranking accuracy on ActivityNet-FOIL dataset. The asterisk $*$ means the metric uses video and text as the reference. EMScore$_f$ means the fine-grained EMScore.}
\label{tab9}
\end{table}

EMScore\cite{shi2022emscore} introduced the ActivityNet-FOIL dataset by injecting foil visual concepts into the original captions from ActivityNet Captions ae-test split. It contains 1,900 correct-foil paragraph pairs, and at least one sentence in the foil paragraph contains a foil visual concept. This experiment uses different metrics to evaluate the correct-foil paragraph pairs and compute the pairwise ranking accuracy. The results are shown in Table \ref{tab9}. We can see that when just using video as the reference, FactVC is better than previous metrics. When using both video and text as the reference, FactVC achieves the highest accuracy of 94.3\%.

\subsection{Cross-Dataset Experiments}

We conduct a cross-dataset experiment to test the generalizability of the FactVC metric. We use different datasets to finetune CLIP model and test them on ActivityNet-Fact and YouCook2-Fact datasets. The results are shown in Table \ref{tab10}. Compared to the CLIP model without finetuning, our finetuned method can obviously improve the metric performance. Even training on a different dataset, the FactVC metric still performs well. Considering the huge domain gap between ActivityNet (ANet) and YouCook2 (You2) datasets, our FactVC metric has good generalizability on different video categories and textual styles.

\begin{table}[htbp]
\begin{center}
\begin{tabular}{l|c|ccc}
\toprule
\multicolumn{5}{c}{ActivityNet-Fact Dataset}\\
\midrule
Finetune Data & ref & Para & Sent & Word\\
\midrule
None & \multirow{4}{*}{VT} & 0.512 & 0.433 & 0.510\\
ANet & & 0.545 & 0.460 & 0.544\\
You2 & & 0.540 & 0.457 & 0.533\\
ANet + You2 & & \textbf{0.551} & \textbf{0.465} & \textbf{0.545}\\
\midrule
\multicolumn{5}{c}{YouCook2-Fact Dataset}\\
\midrule
Finetune Dataset & ref & Para & Sent & Word\\
\midrule
None & \multirow{4}{*}{VT} & 0.572 & 0.555 & 0.587\\
ANet & & 0.584 & 0.569 & 0.604\\
You2 & & \textbf{0.611} & 0.582 & 0.613\\
ANet + You2 & & 0.606 & \textbf{0.583} & \textbf{0.615}\\
\bottomrule
\end{tabular}
\end{center}
\caption{FactVC metric cross-dataset experiments. We show the Pearson correlation between metrics and human annotation, using input video and human-written caption (VT) as reference.}
\label{tab10}
\end{table}

For more implementation details, experiments and qualitative analysis, please refer to Appendix \ref{sec:extra_exp} and \ref{sec:qualitative}.

\subsection{Transferring to Image Captioning}

\begin{table}[htbp]
\begin{center}
\begin{tabular}{l|c|ccc}
\toprule
Metric & ref & Likert & Binary & Word\\
\midrule
Bleu4 & T & 0.266 & 0.230 & 0.252\\
METEOR & T & 0.308 & 0.239 & 0.295\\
Rouge-L & T & 0.364 & 0.289 & 0.361\\
CIDEr & T & 0.375 & 0.300 & 0.336\\
\midrule
CLIPScore & V & 0.359 & 0.220 & 0.297\\
RefCLIPScore & VT & 0.457 & 0.298 & 0.398\\
CLIPScore$^{*}$ & V & 0.364 & 0.238 & 0.309\\
RefCLIPScore$^{*}$ & VT & 0.466 & 0.315 & 0.409\\
CLIPScore$^{**}$ & V & 0.398 & 0.249 & 0.367\\
RefCLIPScore$^{**}$ & VT & \textbf{0.513} & \textbf{0.341} & \textbf{0.478}\\
\bottomrule
\end{tabular}
\end{center}
\caption{Pearson correlation between automatic metrics and human annotation on MSCOCO-Fact dataset. $*$ and $**$ mean using video-finetuned and image-finetuned CLIP model respectively.}
\label{tab11}
\end{table}

According to the above experiments, FactVC performs well in evaluating the factuality of video captioning. We want to know whether our method can transfer to image captioning. So we additionally collect an annotated factuality dataset MSCOCO-Fact based on 200 MSCOCO\cite{lin2014microsoft} test images and five recent image captioning models' outputs. Unlike video captioning where a caption is a multi-sentence paragraph, an image caption is a single sentence. We collect three kinds of factuality annotation for each image caption: Likert (1-5 factuality score), Binary (0 or 1, indicating whether the sentence has a factual error), Word (whether each word has a factual error, and we use the ratio of factual words as word-level annotation score).

We test the correlation between image captioning metrics and human factuality annotation. Results are shown in Table \ref{tab11}. CLIPScore and RefCLIPScore\cite{hessel2021clipscore} are CLIP-based metrics for image captioning. CLIPScore$^{*}$ and RefCLIPScore$^{*}$ use our video-finetuned CLIP model. We further construct a training set from image caption data using the same text augmentation skills, and finetune the CLIP model to get CLIPScore$^{**}$ and RefCLIPScore$^{**}$. From Table \ref{tab11}, we can see that with our finetuned method, CLIPScore and RefCLIPScore can better measure the factuality of image captions. More details of the MSCOCO-Fact dataset are in Appendix \ref{sec:mscoco}.

\section{Conclusion}
In this work, we focus on the factuality evaluation in video captioning. We first collect two human-annotated factuality datasets for video captioning and find that hallucination is a severe problem in video captioning, with 57.0\% of the model-generated sentences having different kinds of factual errors. However, most existing metrics show little correlation with human annotation. So we propose a new factuality metric FactVC. It is trained on an automatically-constructed training set and correlates much better with the factuality annotation. Experiments also show the potential of our method in evaluating image captions. Although factual consistency is a hot research topic in text-to-text tasks, it is less studied in video captioning. We hope our work can fill this research gap and promote further research in video captioning.

{\small
\bibliographystyle{ieee_fullname}
\bibliography{egbib}

\begin{thebibliography}{10}\itemsep=-1pt

\bibitem{anderson2018bottom}
Peter Anderson, Xiaodong He, Chris Buehler, Damien Teney, Mark Johnson, Stephen
  Gould, and Lei Zhang.
\newblock Bottom-up and top-down attention for image captioning and visual
  question answering.
\newblock In {\em Proceedings of the IEEE conference on computer vision and
  pattern recognition}, pages 6077--6086, 2018.

\bibitem{banerjee2005meteor}
Satanjeev Banerjee and Alon Lavie.
\newblock Meteor: An automatic metric for mt evaluation with improved
  correlation with human judgments.
\newblock In {\em Proceedings of the acl workshop on intrinsic and extrinsic
  evaluation measures for machine translation and/or summarization}, pages
  65--72, 2005.

\bibitem{bang2023multitask}
Yejin Bang, Samuel Cahyawijaya, Nayeon Lee, Wenliang Dai, Dan Su, Bryan Wilie,
  Holy Lovenia, Ziwei Ji, Tiezheng Yu, Willy Chung, et~al.
\newblock A multitask, multilingual, multimodal evaluation of chatgpt on
  reasoning, hallucination, and interactivity.
\newblock {\em arXiv preprint arXiv:2302.04023}, 2023.

\bibitem{devaraj2022evaluating}
Ashwin Devaraj, William Sheffield, Byron~C Wallace, and Junyi~Jessy Li.
\newblock Evaluating factuality in text simplification.
\newblock In {\em Proceedings of the 60th Annual Meeting of the Association for
  Computational Linguistics (Volume 1: Long Papers)}, pages 7331--7345, 2022.

\bibitem{durmus2020feqa}
Esin Durmus, He He, and Mona Diab.
\newblock Feqa: A question answering evaluation framework for faithfulness
  assessment in abstractive summarization.
\newblock In {\em Proceedings of the 58th Annual Meeting of the Association for
  Computational Linguistics}, pages 5055--5070, 2020.

\bibitem{falke2019ranking}
Tobias Falke, Leonardo~FR Ribeiro, Prasetya~Ajie Utama, Ido Dagan, and Iryna
  Gurevych.
\newblock Ranking generated summaries by correctness: An interesting but
  challenging application for natural language inference.
\newblock In {\em Proceedings of the 57th Annual Meeting of the Association for
  Computational Linguistics}, pages 2214--2220, 2019.

\bibitem{ging2020coot}
Simon Ging, Mohammadreza Zolfaghari, Hamed Pirsiavash, and Thomas Brox.
\newblock Coot: Cooperative hierarchical transformer for video-text
  representation learning.
\newblock {\em Advances in neural information processing systems},
  33:22605--22618, 2020.

\bibitem{gokhale2022semantically}
Tejas Gokhale, Abhishek Chaudhary, Pratyay Banerjee, Chitta Baral, and Yezhou
  Yang.
\newblock Semantically distributed robust optimization for vision-and-language
  inference.
\newblock In {\em Findings of the Association for Computational Linguistics:
  ACL 2022}, pages 1493--1513, 2022.

\bibitem{hessel2021clipscore}
Jack Hessel, Ari Holtzman, Maxwell Forbes, Ronan Le~Bras, and Yejin Choi.
\newblock Clipscore: A reference-free evaluation metric for image captioning.
\newblock In {\em Proceedings of the 2021 Conference on Empirical Methods in
  Natural Language Processing}, pages 7514--7528, 2021.

\bibitem{honovich2021q2}
Or Honovich, Leshem Choshen, Roee Aharoni, Ella Neeman, Idan Szpektor, and Omri
  Abend.
\newblock Q2:: Evaluating factual consistency in knowledge-grounded dialogues
  via question generation and question answering.
\newblock In {\em Proceedings of the 2021 Conference on Empirical Methods in
  Natural Language Processing}, pages 7856--7870, 2021.

\bibitem{karpathy2015deep}
Andrej Karpathy and Li Fei-Fei.
\newblock Deep visual-semantic alignments for generating image descriptions.
\newblock In {\em Proceedings of the IEEE conference on computer vision and
  pattern recognition}, pages 3128--3137, 2015.

\bibitem{krippendorff2011computing}
Klaus Krippendorff.
\newblock Computing krippendorff's alpha-reliability.
\newblock 2011.

\bibitem{krishna2017dense}
Ranjay Krishna, Kenji Hata, Frederic Ren, Li Fei-Fei, and Juan Carlos~Niebles.
\newblock Dense-captioning events in videos.
\newblock In {\em Proceedings of the IEEE international conference on computer
  vision}, pages 706--715, 2017.

\bibitem{kryscinski2020evaluating}
Wojciech Kry{\'s}ci{\'n}ski, Bryan McCann, Caiming Xiong, and Richard Socher.
\newblock Evaluating the factual consistency of abstractive text summarization.
\newblock In {\em Proceedings of the 2020 Conference on Empirical Methods in
  Natural Language Processing (EMNLP)}, pages 9332--9346, 2020.

\bibitem{lei2020mart}
Jie Lei, Liwei Wang, Yelong Shen, Dong Yu, Tamara Berg, and Mohit Bansal.
\newblock Mart: Memory-augmented recurrent transformer for coherent video
  paragraph captioning.
\newblock In {\em Proceedings of the 58th Annual Meeting of the Association for
  Computational Linguistics}, pages 2603--2614, 2020.

\bibitem{lin2004rouge}
Chin-Yew Lin.
\newblock Rouge: A package for automatic evaluation of summaries.
\newblock In {\em Text summarization branches out}, pages 74--81, 2004.

\bibitem{lin2014microsoft}
Tsung-Yi Lin, Michael Maire, Serge Belongie, James Hays, Pietro Perona, Deva
  Ramanan, Piotr Doll{\'a}r, and C~Lawrence Zitnick.
\newblock Microsoft coco: Common objects in context.
\newblock In {\em European conference on computer vision}, pages 740--755.
  Springer, 2014.

\bibitem{luo2020univl}
Huaishao Luo, Lei Ji, Botian Shi, Haoyang Huang, Nan Duan, Tianrui Li, Jason
  Li, Taroon Bharti, and Ming Zhou.
\newblock Univl: A unified video and language pre-training model for multimodal
  understanding and generation.
\newblock {\em arXiv preprint arXiv:2002.06353}, 2020.

\bibitem{maynez2020faithfulness}
Joshua Maynez, Shashi Narayan, Bernd Bohnet, and Ryan McDonald.
\newblock On faithfulness and factuality in abstractive summarization.
\newblock In {\em Proceedings of the 58th Annual Meeting of the Association for
  Computational Linguistics}, pages 1906--1919, 2020.

\bibitem{miech2019howto100m}
Antoine Miech, Dimitri Zhukov, Jean-Baptiste Alayrac, Makarand Tapaswi, Ivan
  Laptev, and Josef Sivic.
\newblock Howto100m: Learning a text-video embedding by watching hundred
  million narrated video clips.
\newblock In {\em Proceedings of the IEEE/CVF International Conference on
  Computer Vision}, pages 2630--2640, 2019.

\bibitem{narayan2018don}
Shashi Narayan, Shay~B Cohen, and Mirella Lapata.
\newblock Don’t give me the details, just the summary! topic-aware
  convolutional neural networks for extreme summarization.
\newblock In {\em Proceedings of the 2018 Conference on Empirical Methods in
  Natural Language Processing}, pages 1797--1807, 2018.

\bibitem{pagnoni2021understanding}
Artidoro Pagnoni, Vidhisha Balachandran, and Yulia Tsvetkov.
\newblock Understanding factuality in abstractive summarization with frank: A
  benchmark for factuality metrics.
\newblock In {\em Proceedings of the 2021 Conference of the North American
  Chapter of the Association for Computational Linguistics: Human Language
  Technologies}, pages 4812--4829, 2021.

\bibitem{pan2020spatio}
Boxiao Pan, Haoye Cai, De-An Huang, Kuan-Hui Lee, Adrien Gaidon, Ehsan Adeli,
  and Juan~Carlos Niebles.
\newblock Spatio-temporal graph for video captioning with knowledge
  distillation.
\newblock In {\em Proceedings of the IEEE/CVF Conference on Computer Vision and
  Pattern Recognition}, pages 10870--10879, 2020.

\bibitem{papineni2002bleu}
Kishore Papineni, Salim Roukos, Todd Ward, and Wei-Jing Zhu.
\newblock Bleu: a method for automatic evaluation of machine translation.
\newblock In {\em Proceedings of the 40th annual meeting of the Association for
  Computational Linguistics}, pages 311--318, 2002.

\bibitem{radford2021learning}
Alec Radford, Jong~Wook Kim, Chris Hallacy, Aditya Ramesh, Gabriel Goh,
  Sandhini Agarwal, Girish Sastry, Amanda Askell, Pamela Mishkin, Jack Clark,
  et~al.
\newblock Learning transferable visual models from natural language
  supervision.
\newblock In {\em International Conference on Machine Learning}, pages
  8748--8763. PMLR, 2021.

\bibitem{rennie2017self}
Steven~J Rennie, Etienne Marcheret, Youssef Mroueh, Jerret Ross, and Vaibhava
  Goel.
\newblock Self-critical sequence training for image captioning.
\newblock In {\em Proceedings of the IEEE conference on computer vision and
  pattern recognition}, pages 7008--7024, 2017.

\bibitem{rohrbach2018object}
Anna Rohrbach, Lisa~Anne Hendricks, Kaylee Burns, Trevor Darrell, and Kate
  Saenko.
\newblock Object hallucination in image captioning.
\newblock In {\em Proceedings of the 2018 Conference on Empirical Methods in
  Natural Language Processing}, pages 4035--4045, 2018.

\bibitem{shi2022emscore}
Yaya Shi, Xu Yang, Haiyang Xu, Chunfeng Yuan, Bing Li, Weiming Hu, and
  Zheng-Jun Zha.
\newblock Emscore: Evaluating video captioning via coarse-grained and
  fine-grained embedding matching.
\newblock In {\em Proceedings of the IEEE/CVF Conference on Computer Vision and
  Pattern Recognition}, pages 17929--17938, 2022.

\bibitem{song2021towards}
Yuqing Song, Shizhe Chen, and Qin Jin.
\newblock Towards diverse paragraph captioning for untrimmed videos.
\newblock In {\em Proceedings of the IEEE/CVF Conference on Computer Vision and
  Pattern Recognition}, pages 11245--11254, 2021.

\bibitem{vedantam2015cider}
Ramakrishna Vedantam, C Lawrence~Zitnick, and Devi Parikh.
\newblock Cider: Consensus-based image description evaluation.
\newblock In {\em Proceedings of the IEEE conference on computer vision and
  pattern recognition}, pages 4566--4575, 2015.

\bibitem{venugopalan2015sequence}
Subhashini Venugopalan, Marcus Rohrbach, Jeffrey Donahue, Raymond Mooney,
  Trevor Darrell, and Kate Saenko.
\newblock Sequence to sequence-video to text.
\newblock In {\em Proceedings of the IEEE international conference on computer
  vision}, pages 4534--4542, 2015.

\bibitem{wang2020asking}
Alex Wang, Kyunghyun Cho, and Mike Lewis.
\newblock Asking and answering questions to evaluate the factual consistency of
  summaries.
\newblock In {\em Proceedings of the 58th Annual Meeting of the Association for
  Computational Linguistics}, pages 5008--5020, 2020.

\bibitem{wang2022ofa}
Peng Wang, An Yang, Rui Men, Junyang Lin, Shuai Bai, Zhikang Li, Jianxin Ma,
  Chang Zhou, Jingren Zhou, and Hongxia Yang.
\newblock Ofa: Unifying architectures, tasks, and modalities through a simple
  sequence-to-sequence learning framework.
\newblock In {\em International Conference on Machine Learning}, pages
  23318--23340. PMLR, 2022.

\bibitem{wang2021end}
Teng Wang, Ruimao Zhang, Zhichao Lu, Feng Zheng, Ran Cheng, and Ping Luo.
\newblock End-to-end dense video captioning with parallel decoding.
\newblock In {\em Proceedings of the IEEE/CVF International Conference on
  Computer Vision}, pages 6847--6857, 2021.

\bibitem{zhang2021vinvl}
Pengchuan Zhang, Xiujun Li, Xiaowei Hu, Jianwei Yang, Lei Zhang, Lijuan Wang,
  Yejin Choi, and Jianfeng Gao.
\newblock Vinvl: Revisiting visual representations in vision-language models.
\newblock In {\em Proceedings of the IEEE/CVF Conference on Computer Vision and
  Pattern Recognition}, pages 5579--5588, 2021.

\bibitem{zhang2019bertscore}
Tianyi Zhang, Varsha Kishore, Felix Wu, Kilian~Q Weinberger, and Yoav Artzi.
\newblock Bertscore: Evaluating text generation with bert.
\newblock In {\em International Conference on Learning Representations}, 2019.

\bibitem{zhou2018towards}
Luowei Zhou, Chenliang Xu, and Jason~J Corso.
\newblock Towards automatic learning of procedures from web instructional
  videos.
\newblock In {\em Thirty-Second AAAI Conference on Artificial Intelligence},
  2018.

\bibitem{zhou2018end}
Luowei Zhou, Yingbo Zhou, Jason~J Corso, Richard Socher, and Caiming Xiong.
\newblock End-to-end dense video captioning with masked transformer.
\newblock In {\em Proceedings of the IEEE Conference on Computer Vision and
  Pattern Recognition}, pages 8739--8748, 2018.

\end{thebibliography}
}

\clearpage
\appendix
\section{Annotation Protocol and examples}
\label{sec:annotation}

The detailed annotation instructions and protocol we provided to the annotators are shown in Table \ref{tab12}. Two annotation examples are provided in Table \ref{tab13}. 

We paid the annotators 12 dollars per hour, more than the local average minimum wage for human annotation. We checked that all content in the datasets contained no personal information about the annotators.

\begin{table*}[htbp]
    \begin{tabularx}{\textwidth}{X}
    \toprule
    \makecell[c]{\textbf{Annotation Instructions}}\\
    \midrule
    \textbf{About the task:}\\
    Our task is video captioning, which uses AI models to generate text descriptions about the video content automatically. Currently, AI models can generate good captions for short and simple videos. However, while handling more complex videos, the models often generate descriptions with factual errors that contradict the video or describe something not appearing in the video. Your task is to annotate the factual consistency between AI-generated captions and video content.\\
    \midrule
    \textbf{About the annotation method:}\\
    You need to annotate 200(100) videos in total. Each video contains five paragraph captions from five different AI models, and each paragraph contains several sentences. You need to watch each video completely and then label the five captions. For each sentence, you need to judge whether it is fact-consistent with the video content. If the sentence is fact-inconsistent, mark the words/phrases with factual errors. For the whole paragraph, you need to give a factuality Likert scale(1-5, the higher, the better):
    \begin{itemize}
    \setlength\itemsep{0pt}
    \item 5: There are no obvious factual errors.
    \item 4: There are a few minor factual errors. Most parts are fact-consistent.
    \item 3: There are factual errors, but the fact-consistent contents are more.
    \item 2: There are more factual errors, and the fact-inconsistent contents are more.
    \item 1: There are a lot of severe factual errors. It can hardly describe the video content.
    \end{itemize}\\
    \midrule
    \textbf{Other tips:}
    \begin{itemize}
    \setlength\itemsep{0pt}
    \item AI models sometimes generate poor-quality descriptions, which may affect your annotation. If the sentence has minor grammar errors, you need to label it according to the corrected sentence. Otherwise, you need to label it as fact-inconsistent. There is a special word ``UNK'' (unknown word) which you should label as inconsistent.
    \item About the relationship between multiple sentences in a paragraph. The sentences are not in strict time order. The first sentence may describe the second half of the video, while the second may describe the first half. As a result, you should annotate each sentence separately and not consider the relation between multiple sentences.
    \item About the completeness of captions. You should only focus on the captions' correctness and not consider whether the caption describes the video completely. For example, caption 1 only describes a part of the video without factual errors, while caption 2 describes most video content with factual errors. The factuality of caption 1 is better than caption 2.
    \item About the commonsense. You can use commonsense during annotation. For example, if the video content is ``a person laying on the bed with eyes closed'', then the caption ``a person is sleeping on the bed'' is correct.
    \item About the annotation of phrases/words. You should mark as few words as possible if there are factual errors. For example, if the video content is about ``A person lays on the bed'', and the caption is ``A person sits on the bed'', you should mark ``sits''. If the video content is about ``A person lays on the ground'', and the caption is ``A person sits on the bed'', you should mark the whole phrase ``sits on the bed''.
    \item The order of AI models is shuffled. Do not assume the first caption comes from model 1, and the second caption comes from model 2, etc. After completing a video annotation, you should go back and check it and ensure your standard is consistent.
    \item We provide you with several annotation examples. Please read them before starting your annotation. This is very helpful for understanding the annotation method.
    \end{itemize}\\
    \bottomrule
  \end{tabularx}
  \caption{The detailed annotation instructions and protocol that we provided to the annotators.}
  \label{tab12}
\end{table*}

\begin{table*}[htbp]
    \centering
    \begin{subtable}[t]{0.49\textwidth}
    \centering
    \begin{tabularx}{\linewidth}[t]{X}
    \toprule
    Video content:\\
    \includegraphics[width=\linewidth]{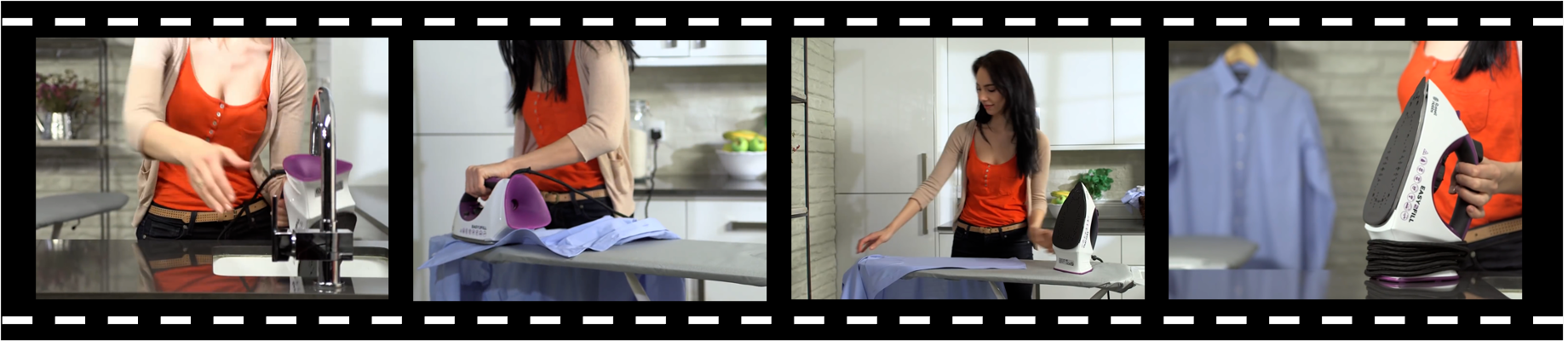}\\ 
    \midrule
    A \textcolor{red}{man} is seen \textcolor{red}{speaking to the camera} and leads into \textcolor{red}{him} holding up \textcolor{red}{a pair of} tools.\\
    The \textcolor{red}{man} then begins ironing the shirt while \textcolor{red}{speaking to the camera}.\\
    \textcolor{red}{He} continues to iron the \textcolor{red}{iron} and ends by showing off the finished product.\\
    Paragraph score: 2\\
    \midrule
    A woman is seen ironing \textcolor{red}{a pair of pants} on an ironing board while \textcolor{red}{speaking to the camera}.\\
    She continues ironing the \textcolor{red}{pants} and ends by showing off the shirt.\\
    Paragraph score: 2\\
    \midrule
    A woman is ironing a shirt on an ironing board.\\
    She shows off \textcolor{red}{a pair of pants}.\\
    She then irons the shirt on the ironing board.\\
    Paragraph score: 4\\
    \midrule
    A woman is seen \textcolor{red}{speaking to the camera} and leads into a large iron of a large iron.\\
    The woman then begins ironing the shirt and irons the \textcolor{red}{iron}.\\
    The woman continues to iron the \textcolor{red}{iron} and shows off the iron.\\
    Paragraph score: 3\\
    \midrule
    She then shows the iron the iron and continues to use the iron.\\
    She then irons the ironing the shirt and begins ironing the \textcolor{red}{pants}.\\
    A woman is standing in a kitchen \textcolor{red}{talking to the camera}.\\
    Paragraph score: 3\\
    \bottomrule
    \end{tabularx}
    \end{subtable}
    \hspace{\fill}
    \begin{subtable}[t]{0.49\textwidth}
    \centering
    \begin{tabularx}{\linewidth}[t]{X}
    \toprule
    Video content:\\
    \includegraphics[width=\linewidth]{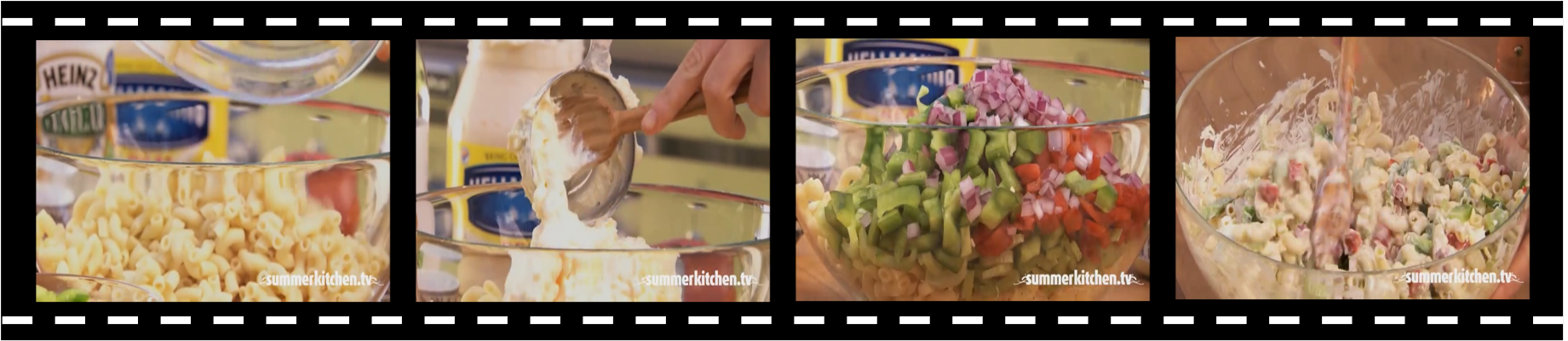}\\ 
    \midrule
    Add \textcolor{red}{chickpeas lemon juice} and \textcolor{red}{lemon juice} to a bowl.\\
    Add \textcolor{red}{flour} salt pepper and a spoon of \textcolor{red}{chicken breast} and mix.\\
    Add chopped \textcolor{red}{tomatoes} chopped \textcolor{red}{spring onions} and a little salt and pepper.\\
    Toss and mix everything together.\\
    Paragraph score: 2\\
    \midrule
    Pour macaroni and \textcolor{red}{milk} on the pasta.\\
    Add pasta sugar salt pepper and vinegar to the salad.\\
    Add salt and pepper and mix.\\
    Mix the salad.\\
    Paragraph score: 4\\
    \midrule
    Pour boiled macaroni and boiled macaroni in a bowl.\\
    Add some mayonnaise and blend until smooth.\\
    Add diced celery and minced \textcolor{red}{garlic} to a bowl.\\
    Mix everything together.\\
    Paragraph score: 4\\
    \midrule
    Add 1 cup of chopped \textcolor{red}{green onions} and 1 cup of chopped \textcolor{red}{green onions}.\\
    Plate the \textcolor{red}{meat} with the \textcolor{red}{sauce} and \textcolor{red}{bread crumbs}.\\
    Add diced onion celery celery and \textcolor{red}{mint} to the \textcolor{red}{food processor}.\\
    Mix the ingredients in the bowl.\\
    Paragraph score: 2\\
    \midrule
    Add pasta to a bowl.\\
    Mix mayonnaise mayonnaise mayonnaise salt and pepper.\\
    Add the \textcolor{red}{cabbage} celery and red bell pepper to the \textcolor{red}{cabbage}.\\
    Toss the salad.\\
    Paragraph score: 4\\
    \bottomrule
    \end{tabularx}
    \end{subtable}
     \caption{Annotation examples. For each example, we show the video content, five paragraph captions, and the paragraph factuality scores. The phrases/words that are not factual are marked in red.}
     \label{tab13}
\end{table*}

\section{Training Dataset Generation}
\label{sec:data_gen}
A detailed description of the data generation process is shown in Algorithm \ref{alg:data_gen}.

\begin{algorithm}
\caption{The algorithm to generate dataset}\label{alg:data_gen}
\begin{algorithmic}
\Require\\ 
$S$ - set of videos $V$ and captions $T$\\
$\mathcal{T}^{+}$ - set of positive transformations\\
$\mathcal{T}^{-}$ - set of negative transformations\\
\Function{GENERATE\_DATA}{$S, \mathcal{T}^{+}, \mathcal{T}^{-}$}
\State $\mathcal{P} \gets \varnothing$  \Comment{set of positive data}
\For{$(V, T)$ \textbf{in} $S$}
\State $\mathcal{P} \gets \mathcal{P} \cap \{(V, T)\}$
\For{$fn$ \textbf{in} $\mathcal{T}^{+}$}
\State $T^+ \gets fn(T)$
\State $\mathcal{P} \gets \mathcal{P} \cap \{(V, T^+)\}$
\EndFor
\EndFor
\State $\mathcal{D} \gets \varnothing$  \Comment{set of data pairs}
\For{$(V, T^+)$ \textbf{in} $\mathcal{P}$}
\For{$fn$ \textbf{in} $\mathcal{T}^{-}$}
\State $T^- \gets fn(T^+)$
\State $\mathcal{D} \gets \mathcal{D} \cap \{(V, T^+, T^-)\}$
\EndFor
\EndFor
\EndFunction

\State \Return $\mathcal{D}$
\end{algorithmic}
\end{algorithm}

\begin{table*}[ht]
\centering
\begin{tabular}{l|ccc|ccc|ccc}
\toprule
\multirow{2}{*}{\makecell[c]{FactVC\\$\lambda$ value}} & \multicolumn{3}{c}{Video as ref} & \multicolumn{3}{|c}{Text as ref} & \multicolumn{3}{|c}{Video \& Text as ref}\\
\cmidrule{2-10}
& Para & Sent & Word & Para & Sent & Word & Para & Sent & Word\\
\midrule
$\lambda = 0.0$ & 0.421 & 0.343 & 0.448 & 0.508 & 0.435 & 0.490 & 0.537 & 0.455 & 0.528\\
$\lambda = 0.1$ & \textbf{0.444} & \textbf{0.363} & 0.460 & \textbf{0.515} & 0.441 & \textbf{0.502} & \textbf{0.547} & \textbf{0.465} & \textbf{0.541}\\
$\lambda = 0.2$ & 0.438 & 0.355 & 0.462 & 0.514 & \textbf{0.443} & 0.499 & 0.543 & 0.461 & 0.536\\
$\lambda = 0.3$ & 0.436 & 0.353 & \textbf{0.463} & 0.514 & \textbf{0.443} & 0.500 & 0.541 & 0.460 & 0.536\\
$\lambda = 0.5$ & 0.427 & 0.346 & 0.459 & 0.513 & 0.442 & 0.500 & 0.537 & 0.457 & 0.534\\
$\lambda = 1.0$ & 0.402 & 0.325 & 0.439 & 0.509 & 0.439 & 0.499 & 0.525 & 0.446 & 0.527\\
\bottomrule
\end{tabular}
\caption{Pearson correlation between FactVC(using different $\lambda$ value) and human factuality annotation on ActivityNet-Fact. We sample one frame from each video clip in this experiment. The best performance in each column is marked in bold.}
\label{tab14}
\end{table*}

\begin{table*}[htbp]
\centering
\begin{tabular}{l|c|ccc|ccc|ccc}
\toprule
\multirow{2}{*}{CLIP Model} & \multirow{2}{*}{Size} & \multicolumn{3}{c}{Video as ref} & \multicolumn{3}{|c}{Text as ref} & \multicolumn{3}{|c}{Video \& Text as ref}\\
\cmidrule{3-11}
 & & Para & Sent & Word & Para & Sent & Word & Para & Sent & Word\\
\midrule
RN50 & 102M & \textbf{0.317} & \textbf{0.265} & \textbf{0.357} & 0.478 & 0.412 & 0.480 & 0.488 & 0.418 & 0.497\\
RN101 & 120M & 0.289 & 0.239 & 0.316 & 0.486 & 0.410 & 0.491 & 0.490 & 0.412 & 0.500\\
RN50x4 & 178M & 0.282 & 0.230 & 0.328 & 0.478 & 0.403 & 0.497 & 0.480 & 0.403 & 0.507\\
RN50x16 & 291M & 0.296 & 0.230 & 0.333 & \textbf{0.503} & \textbf{0.435} & \textbf{0.510} & \textbf{0.502} & \textbf{0.428} & \textbf{0.517}\\
RN50x64 & 623M & 0.276 & 0.216 & 0.320 & 0.470 & 0.401 & 0.466 & 0.467 & 0.394 & 0.474\\
\midrule
ViT-B/32 & 151M & 0.321 & 0.253 & 0.369 & 0.472 & 0.396 & 0.466 & 0.483 & 0.402 & 0.489\\
ViT-B/16 & 150M & \textbf{0.349} & \textbf{0.281} & \textbf{0.388} & \textbf{0.497} & \textbf{0.425} & \textbf{0.483} & \textbf{0.512} & \textbf{0.433} & \textbf{0.510}\\
ViT-L/14 & 428M & 0.332 & 0.254 & 0.356 & 0.456 & 0.401 & 0.442 & 0.469 & 0.404 & 0.462\\
ViT-L/14-336px & 428M & 0.340 & 0.268 & 0.364 & 0.454 & 0.400 & 0.441 & 0.468 & 0.405 & 0.463\\
\bottomrule
\end{tabular}
\caption{Pearson correlation between FactVC (using different CLIP models, no finetuning) and human factuality annotation on ActivityNet-Fact. We show each model's size (number of parameters). The best ResNet(RN) CLIP model and Vision-Transformer(ViT) CLIP model are marked in bold.}
\label{tab15}
\end{table*}

\section{Experiments}
\label{sec:extra_exp}

\subsection{Implementation details}
We use the training split of ActivityNet Captions and YouCook2 to construct our training and validation set. The training and validation set size are 44,820 and 5,180 for ActivityNet and 18,029 and 1,971 for YouCook2. For CLIP finetuning, we start with the pretrained ViT-B/16 CLIP model. We sample three frames from each video clip uniformly. We set the margin $M$ in Eq (3) to 5.0 and the loss weight $\lambda$ in Eq (4) to 0.1. We finetune the projection layers of the CLIP model for three epochs with a batch size of 256 and learning rate of $5e-5$. During score calculation, we set the balance factor $\alpha$ in Eq (5) to 0.75, favoring fine-grained scores more than coarse-grained ones. Regarding the complexity cost, we finetune CLIP 3 epochs, which cost 4-6 hours on a single 2080Ti GPU card. We will keep the above settings unless otherwise stated. 

\subsection{Extra ablation test}
We explore the impact of the $\lambda$ value in Eq (4). The results are shown in Table \ref{tab14}. Note that when $\lambda = 0$, we only use $\mathcal{L}_{coarse}$ to finetune the CLIP model. From the table, FactVC performs better when choosing a $\lambda$ between $[0.1, 0.3]$. It makes the CLIP model make use of both $\mathcal{L}_{coarse}$ and $\mathcal{L}_{fine}$.

Considering our FactVC is based on the image-text pretrained model CLIP\cite{radford2021learning}, we want to explore the impact of using different CLIP models. The results are shown in Table \ref{tab15}. We test FactVC performance on ActivityNet-Fact with different CLIP models without finetuning in this experiment. The table shows that among ResNet-based CLIP models, RN50 and RN50x16 perform best; among ViT-based CLIP models, ViT-B/16 performs best. This leads us to the conclusion that a larger CLIP model does not necessarily perform better on factuality evaluation. As a result, we use the relatively small ViT-B/16 CLIP model in this work.

\subsection{Model Ranking}
Evaluation metrics are often reported at the system level to compare the performance of different models, and a reliable metric should be consistent with human judgment. We test the performance of the five models on the ActivityNet-Fact dataset using EMScore and FactVC and report the average scores. The results are shown in Table \ref{tab17}. All the metrics are scaled to $[0,1]$. Compared to Sentence-level factuality annotation, EMScore ranks the COOT, PDVC-gt, and PDVC-pred models differently. In contrast, FactVC ranks them consistently with human annotation.

\begin{table}[htbp]
\centering
\begin{tabular}{l|ccc}
\toprule
Models & Annotation & EMScore & FactVC\\
\midrule
MART & 0.413(5) & 0.537(5) & 0.581(5)\\
COOT & 0.446(3) & 0.583\textcolor{red}{(2)} & 0.628(3)\\
PDVC-gt & 0.553(1) & 0.582\textcolor{red}{(3)} & 0.646(1)\\
PDVC-pred & 0.513(2) & 0.592\textcolor{red}{(1)} & 0.644(2)\\
Song & 0.443(4) & 0.573(4) & 0.610(4)\\
\bottomrule
\end{tabular}
\caption{Model performance with ranking on Sentence-level annotation and automatic metrics. We use video and text as the reference to compute EMScore and FactVC metrics. The ranking of each model is shown in parentheses, and the rankings that are inconsistent with human annotation are marked in red.}
\label{tab17}
\end{table}

\begin{table*}[htbp]
    \centering
    \begin{subtable}[t]{\textwidth}
    \centering
    \begin{tabularx}{\linewidth}[t]{X}
    \toprule
    Video 1:\\
    \includegraphics[width=\linewidth]{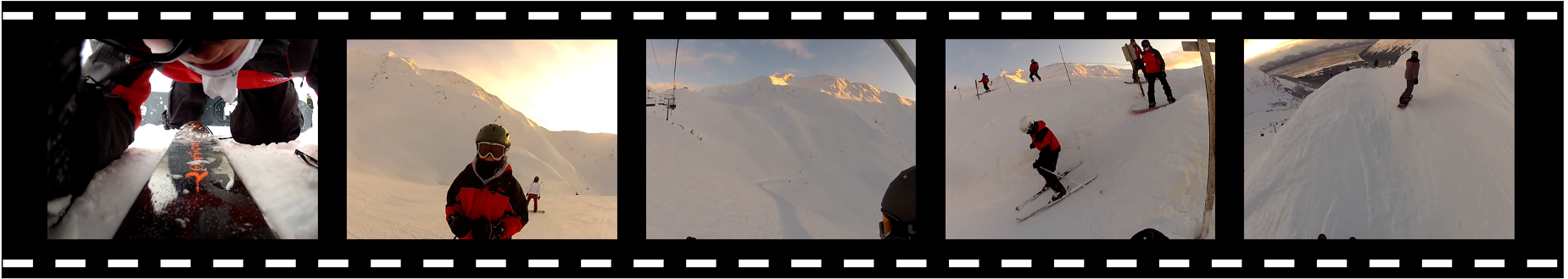}\\ 
    \midrule
    Generated caption:\\
    A person is skiing down a hill of snow. They go over a hill of snow. They continue skiing down the hill together.\\
    \midrule
    Reference captions:\\
    A group of people are on a snowy mountain top.  They are skiing down the numerous hills together.  We see them flip and turn sharply in the driven snow.\\
    A man is crouched down in the snow looking at the camera.  He is then seen skiing through the snow.  He is also seen riding the lifts before skiing again.\\
    \midrule
    Paragraph: 1.0 \quad Sentence: 1.0 \quad Word: 1.0\\
    Bleu2: 0.258 \quad METEOR: 0.320 \quad CIDEr: 0.238 \quad BERTScore: 0.610\\
    EMScore(V): 0.649 \quad EMScore(VT): 0.805 \quad FactVC(V): 0.838 \quad FactVC(VT): 0.905\\
    \bottomrule
    \end{tabularx}
    \end{subtable}
    \hspace{\fill}
    \begin{subtable}[t]{\textwidth}
    \centering
    \begin{tabularx}{\linewidth}[t]{X}
    \toprule
    Video 2:\\
    \includegraphics[width=\linewidth]{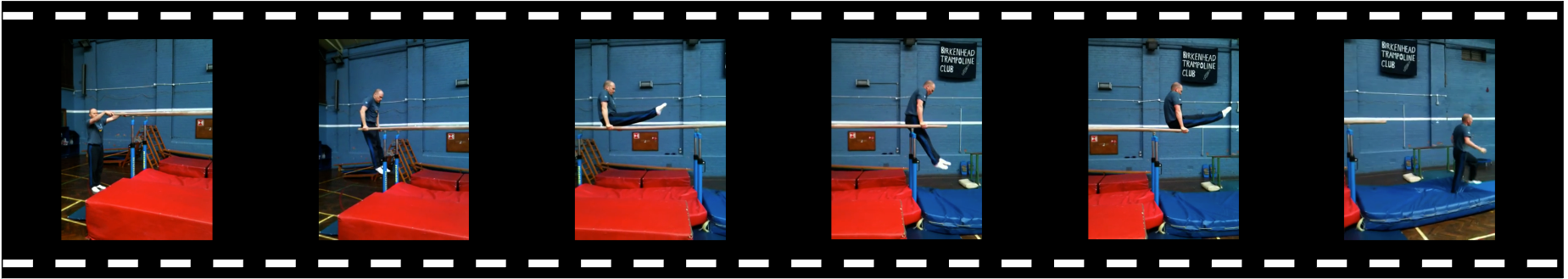}\\ 
    \midrule
    Generated caption:\\
    He does a gymnastics routine on the bars. He does a gymnastics routine on the bars. He dismounts and lands on the \textcolor{red}{bars}.\\
    \midrule
    Reference captions:\\
    A man is seen standing before a set of uneven bars and begins inching himself forward.  He raises his legs up when he stops and continues inches forward.  He moves down all the way to end and jumps off into the mats in the end.\\
    A man stands on front the parallel bars holding it.  The man starts to advance holding on his hands.  The man stops in the middle of the parallel bars, raise his legs and after continues advancing.  Then, the man stops at the end of the bars, again he raises his legs, then exercises up and down.  Next, the man jumps on the mat.\\
    \midrule
    Paragraph: 0.75 \quad Sentence: 0.667 \quad Word: 0.957\\
    Bleu2: 0.144 \quad METEOR: 0.082 \quad CIDEr: 0.0 \quad BERTScore: 0.243\\
    EMScore(V): 0.613 \quad EMScore(VT): 0.582 \quad FactVC(V): 0.711 \quad FactVC(VT): 0.671\\
    \bottomrule
    \end{tabularx}
    \end{subtable}
     \caption{Video captioning evaluation examples. For each example, we show the three-level factuality annotation scores and different metric scores. All scores are scaled in $[0,1]$. For EMScore and FactVC, 'V' means using video as reference and 'VT' means using video and text as the reference. The factual errors are marked in red in the generated captions.}
     \label{tab18}
\end{table*}

\begin{table*}[htbp]
    \centering
    \begin{subtable}[t]{\textwidth}
    \centering
    \begin{tabularx}{\linewidth}[t]{X}
    \toprule
    Video 3:\\
    \includegraphics[width=\linewidth]{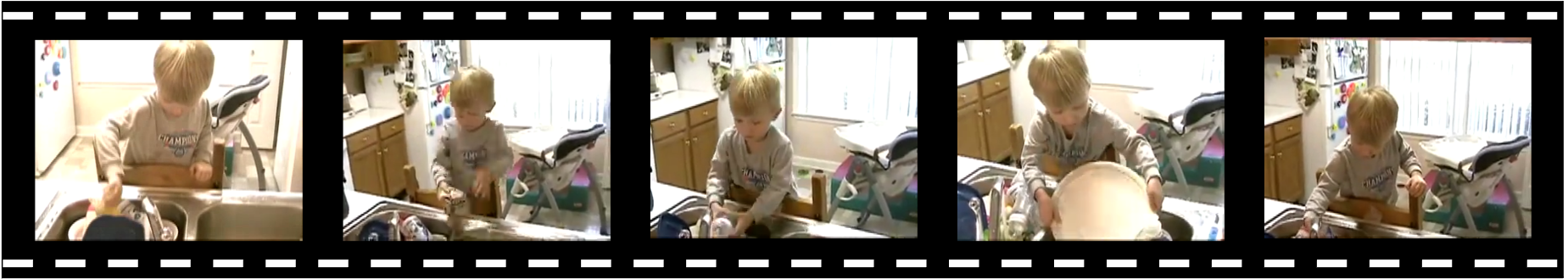}\\ 
    \midrule
    Generated caption:\\
    A young boy is seen \textcolor{red}{sitting} behind a sink with \textcolor{red}{a woman standing behind him}. The boy then begins washing dishes while \textcolor{red}{the boy watches from the side}. The boy continues to \textcolor{red}{brush his face} and ends by turning off the camera.\\
    \midrule
    Reference captions:\\
    A toddler washes dishes in a sink while stand on a chair.  The boy washes a cup, a sip cup and a dish.  After, the boy jumps on the chair and then takes the dish again.\\
    A small child is seen standing before a sink washing dishes.  He wipes around the sink and continues washing dishes.  He puts the clean dishes next to him.\\
    \midrule
    Paragraph: 0.25 \quad Sentence: 0.0 \quad Word: 0.643\\
    Bleu2: 0.379 \quad METEOR: 0.448 \quad CIDEr: 0.121 \quad BERTScore: 0.724\\
    EMScore(V): 0.723 \quad EMScore(VT): 0.630 \quad FactVC(V): 0.497 \quad FactVC(VT): 0.522\\
    \bottomrule
    \end{tabularx}
    \end{subtable}
    \hspace{\fill}
    \begin{subtable}[t]{\textwidth}
    \centering
    \begin{tabularx}{\linewidth}[t]{X}
    \toprule
    Video 4:\\
    \includegraphics[width=\linewidth]{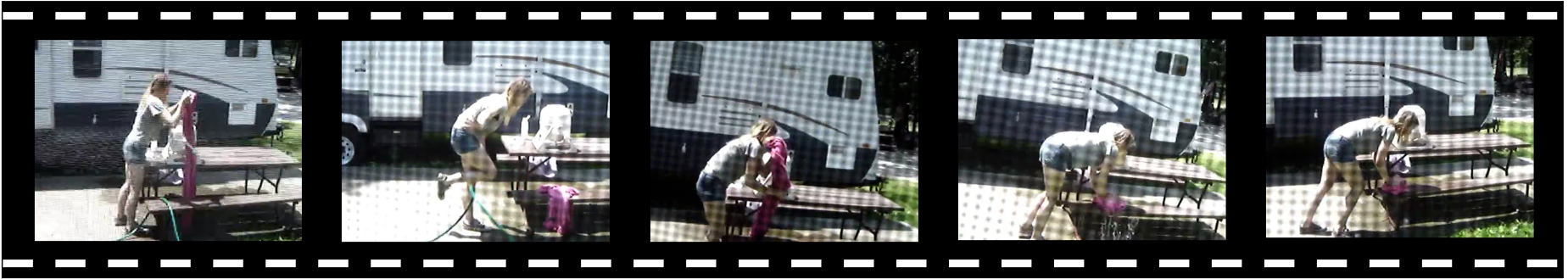}\\ 
    \midrule
    Generated caption:\\
    A \textcolor{red}{man} is seen standing \textcolor{red}{on a track with a stick}. The \textcolor{red}{man} then begins \textcolor{red}{playing with another person on a rope}. The \textcolor{red}{man} continues \textcolor{red}{playing and walks away}.\\
    \midrule
    Reference captions:\\
    A woman is standing at a picnic table outside an RV.  She is using water from a dispenser, pouring it onto material.  She cleans the pink item with the water.\\
    A lady is outside wringing a cloth on a bench.  The lady places the pink cloth down on the bench.  The lady removes a green hose from the brown bench.  The lady press the cloth down on the bench and water drains.\\
    \midrule
    Paragraph: 0.0 \quad Sentence: 0.0 \quad Word: 0.310\\
    Bleu2: 0.180 \quad METEOR: 0.072 \quad CIDEr: 0.008 \quad BERTScore: 0.373\\
    EMScore(V): 0.152 \quad EMScore(VT): 0.129 \quad FactVC(V): 0.033 \quad FactVC(VT): 0.072\\
    \bottomrule
    \end{tabularx}
    \end{subtable}
     \caption{Continued with Table \ref{tab6}. Two more video captioning evaluation examples.}
     \label{tab19}
\end{table*}

\section{FactVC Qualitative Analysis}
\label{sec:qualitative}
We show several video captioning evaluation examples in Table \ref{tab18} and \ref{tab19}. The annotation scores and metric scores are scaled in $[0,1]$ for comparison. In video 1, the generated caption has no obvious factual errors. BERTScore, EMScore, and FactVC assign relatively high scores, while FactVC shows the most confidence. In video 2, the generated caption has a minor factual error, but it has a poor overlap with the references. All text-reference-based metrics give low scores, while FactVC gives a more reasonable factual score. In video 3, the generated caption has many different kinds of factual errors. However, probably because of the semantic overlap, BERTScore and EMScore give it high scores. Our FactVC gives a relatively low score. In video 4, the generated caption is full of severe factual errors, and FactVC correctly gives a very low score. The examples show that our FactVC metric performs best in measuring the factuality of video captions.

\section{MSCOCO-Fact Dataset}
\label{sec:mscoco}
In order to test our method on image captioning, we additionally collect a human-annotated factuality dataset MSCOCO-Fact. We sample 200 MSCOCO\cite{lin2014microsoft} images from Karpathy test split\cite{karpathy2015deep}. We select five image captioning models: \textbf{BUTD}\cite{anderson2018bottom}: a popular image captioning model using bottom-up and top-down attention; \textbf{BUTD-sc}\cite{anderson2018bottom}: use self-critical sequence training method\cite{rennie2017self} to train BUTD model; \textbf{VinVL}\cite{zhang2021vinvl}: a large-scale pretrained captioning model; \textbf{OFA-base, OFA-huge}\cite{wang2022ofa}: a unified multimodal pretrained model that achieves the SOTA performance on MSCOCO caption task. We additionally annotate the factuality of human-generated image captions.

We use a similar annotation protocol as video captioning. We collect three kinds of factuality annotation for each image caption sentence: Likert (1-5 factuality score), Binary (0 or 1, indicating whether the sentence has a factual error), Word (whether each word has a factual error). The MSCOCO-Fact dataset contains 1,200 sentences and 11,703 words, among which 29.3\% of the sentences and 6.2\% of the words have factual errors. Compared to the video captioning dataset ActivityNet-Fact and YouCook2-Fact, MSCOCO-Fact suffers less from factual errors.

\begin{table*}[ht]
\centering
\begin{tabular}{l|cccc|ccc}
\toprule
\multirow{2}{*}{Models} & \multicolumn{4}{c}{Automatic Metrics} & \multicolumn{3}{|c}{Factuality Annotation}\\
\cmidrule{2-8}
& Bleu4 & METEOR & Rouge-L & CIDEr & Likert & Binary & Word\\
\midrule
BUTD & 35.70 & 27.83 & 56.64 & 121.26 & 4.19 & 0.655 & 0.907\\
BUTD-sc & 37.90 & 28.21 & 57.94 & 132.62 & 4.02 & 0.560 & 0.887\\
VinVL & 41.34 & 31.65 & 61.09 & 152.15 & \underline{4.52} & \underline{0.725} & \underline{0.953}\\
OFA-base & \underline{43.73} & \underline{32.01} & \underline{62.03} & \underline{161.95} & 4.50 & \underline{0.725} & 0.951\\
OFA-huge & \textbf{44.55} & \textbf{32.46} & \textbf{63.37} & \textbf{165.86} & \textbf{4.67} & \textbf{0.795} & \textbf{0.969}\\
\midrule
Human & - & - & - & - & 4.64 & 0.780 & 0.963\\
\bottomrule
\end{tabular}
\caption{Image caption performance on automatic metrics and Factuality Annotation on the MSCOCO-Fact dataset. For factuality annotation, we report the average factuality likert score (Likert) and the ratio of factual sentences (Binary) and words (Word). The two highest-rated models are in bold and underlined respectively.}
\label{tab16}
\end{table*}

We evaluate the selected models using both automatic metrics and factuality annotation. The results are shown in Table \ref{tab16}. The SOTA image captioning model OFA-huge performs best, and its factuality even outperforms human-written captions. The pretrained models (OFA, VinVL) show an obvious advantage over non-pretraining models (BUTD). Self-critical sequence training method\cite{rennie2017self} can improve the automatic metrics but may harm the captions' factuality.

\end{document}